\documentclass[letterpaper,journal]{IEEEtran}
\usepackage{amsmath,amsfonts}
\usepackage{algorithmic}
\usepackage{algorithm}
\usepackage{array}
\usepackage[caption=false,font=normalsize,labelfont=sf,textfont=sf]{subfig}
\usepackage{textcomp}
\usepackage{stfloats}
\usepackage{url}
\usepackage{verbatim}
\usepackage{graphicx}
\usepackage{cite}
\usepackage{xcolor}
\usepackage{booktabs}
\usepackage{multirow}
\usepackage{amssymb}
\definecolor{linkblue}{RGB}{0, 0, 255}
\usepackage[
    colorlinks=true,
    allcolors=linkblue
]{hyperref}

\begin{document}

\title{Seeing the Invisible: Physics-Guided Visual Prompting for Temperature- and Radiation-Aware VLA Navigation}

\author{Hojoon Son, Fan Zhang
\thanks{The authors are with the George W. Woodruff School of Mechanical Engineering, Georgia Institute of Technology, Atlanta, GA 30332-0405 USA (e-mail: {\tt\small hojoon.son@gatech.edu}; {\tt\small fan@gatech.edu}).}%
\thanks{This work has been submitted to the IEEE for possible publication. Copyright may be transferred without notice, after which this version may no longer be accessible.}
}

\markboth{}%
{}

\maketitle

\begin{abstract}

Vision-Language-Action (VLA) models have become a major paradigm for Vision-and-Language Navigation (VLN). However, in safety-critical facilities, invisible risks such as radiation or temperature spikes cannot be detected by an RGB camera, and handling each risk is expensive, requiring a new encoder, new data, and model retraining. We propose Physics-Guided Visual Prompting (\textit{PG-VP}), a plug-and-play multimodal perception module that instead reuses what a frozen VLA model already does well: avoiding visible obstacles. Given a proximal radiation or thermal source, PG-VP performs a physics-guided risk assessment to determine the avoidance direction and overlays a corresponding virtual obstacle that moves across consecutive frames (\textit{Dynamic Visual Prompting}). The navigation policy then naturally detours around this invisible hazard. The identical virtual obstacle is used regardless of hazard type, so the visual prompting pattern remains fixed as sensors are added. When no hazard is detected, nothing is rendered, and the policy behaves exactly as it would without PG-VP. We evaluate PG-VP on OmniNav using the val-unseen splits of R2R-CE and RxR-CE, where it guides the policy toward intended low-risk actions in 84.9\% and 83.2\% of cases, at a cost of 6.8 and 7.9 percentage points in navigation success rate. We further test it with distinct scenarios on a real robot in the presence of actual thermal and radiation sources, all without any retraining. The real test shows that PG-VP effectively avoids these invisible hazards, improving worst-10\% average trajectory safety by 63.45\% and 32.59\% against thermal and radiation sources, respectively.
\end{abstract}


\section{INTRODUCTION}

Vision-and-Language Navigation (VLN) is a core problem in achieving generalist navigation capabilities guided by natural language instructions. In particular, implementing VLN in safety-critical facilities, such as nuclear power plants (NPPs), is beneficial for their operations and maintenance (O\&M). Son~\textit{et al.}~\cite{son2025use} proposed a systematic catalog of potential robotic applications in NPP O\&M by analyzing historical nuclear incidents using domain-specific large language models (LLMs). Based on their insights, NPP O\&M requires multi-task robots rather than task-specific ones. Therefore, VLN systems with generalist robots can be valuable not only in everyday environments but also in safety-critical domains. 

Vision-Language-Action (VLA) models have recently emerged, leveraging the foundational
capabilities of Vision-Language Models (VLMs) to unify the perception-to-action pipeline into a jointly trained policy that maps multimodal inputs directly to action tokens, and now represent the latest paradigm of VLN~\cite{cheng2025navila,xue2026omninav,zhang2026qwen,chu2026abot}. Still, a key challenge remains for VLA-based navigation in safety-critical environments: visual perception alone is insufficient. Invisible hazards, such as radiation or temperature spikes, can arise during navigation~\cite{son2025use}, but they leave no signature in the visible spectrum, and thus cannot be detected by a standard RGB camera. A straightforward solution would be to extend a VLA model's input modalities, but this is expensive: it requires a new encoder, new paired multimodal-language data, and retraining for every new modality~\cite{huang2025tactile,jones2025beyond,cheng2026omnivtla,wang2026vision}. More critically, the model offers no guarantee that it preserves its original navigation performance in scenarios without such invisible hazards.

We propose the physics-guided visual prompting (PG-VP) module, which guides the navigation trajectory to naturally detour around invisible proximal hazards by leveraging the frozen VLA model's foundational capability to avoid visible obstacles. It assesses the risk posed by proximal radiation or thermal sources, identifies the avoidance direction, and overlays the corresponding virtual obstacle onto the VLA model's input images. Instead of overlaying a static obstacle onto the scene, PG-VP shifts the virtual obstacle across frames to smoothly guide the trajectory, which we call dynamic visual prompting. As a result, the VLA model treats the injected context as a real obstacle and implicitly avoids the invisible hazards. The main benefits of PG-VP are as follows:
\begin{enumerate}
    \item PG-VP is a plug-and-play visual prompting module that injects non-geometric multimodal physical context into a VLA-based navigation framework in a zero-shot manner, without touching the model weights. When no hazard is detected, it renders no virtual obstacles, fully preserving the native navigation performance. 
    \item PG-VP requires only a risk assessment model per modality. Building this model is far cheaper than adding an encoder, collecting new data, and retraining.
    \item PG-VP renders the same virtual obstacle regardless of modality type. Because the model interprets a single visual prompting pattern rather than a different one for each modality, the alignment with the training distribution remains consistent across sensing modalities.
\end{enumerate}
We used OmniNav as the underlying navigation policy~\cite{xue2026omninav}, evaluated PG-VP's trajectory-guidance effectiveness across unseen R2R-CE and RxR-CE splits~\cite{Matterport3D,anderson2018vision,ku2020room,krantz_vlnce_2020}, and demonstrated physical deployment under actual thermal and radiation sources.

\section{Related Work}
\subsection{VLA-based Robot Navigation}
NaVid~\cite{zhang2024navid} was an early attempt to leverage video-based VLMs, directly mapping natural language instructions to low-level navigation actions from an RGB stream; Uni-NaVid~\cite{zhang2024uni} and NaVILA~\cite{cheng2025navila} established this formulation as a VLA paradigm for navigation. Since then, this has advanced along several directions: broadening generalization across embodiments, tasks, and goal modalities through large-scale training~\cite{zhang2026embodied,hirose2025omnivla,chu2026abot}; compressing accumulated video context and improving systemic efficiency for real-time deployment~\cite{wei2025streamvln,zhang2025mapnav,zheng2025efficient}; injecting geometric priors to recover 3D spatial understanding from RGB input alone~\cite{zeng2026janusvln,liu2026span}; decoupling high-level semantic reasoning from low-level control in dual-system architectures~\cite{wei2026ground,internvla-n1,huang2026sedualvln}; and post-training the policy with self-correction data flywheels~\cite{yu2026correctnav}. At the intersection of these directions, OmniNav~\cite{xue2026omninav} adopts a fast-slow design: a fast system generates short-horizon continuous-space waypoints via a flow-matching policy, while a slow system plans long-horizon sub-goals from frontiers and visual memory. We adopt OmniNav's fast system as our underlying vision-only navigation policy.

\subsection{Extending Perception Beyond RGB}
In VLA-based manipulation, depth can be reconstructed from RGB alone, but contact force, sound, and temperature leave no signature in the visible spectrum. To include these physical quantities in the VLA policy input, researchers must add an additional sensor, and several studies have pursued this approach~\cite{huang2025tactile,cheng2026omnivtla,jones2025beyond,guo2025omnivla}. Yet, adding a new sensor is costly. A new modality requires collecting additional robot data~\cite{wang2026vision} and retraining the model to recover the pretrained alignment~\cite{li2026pointvla,sun2025geovla}. Furthermore, for deployment, platforms without the additional sensor require a separate policy~\cite{zhang2026core}.

In VLA-based navigation, geometric quantities, such as goal coordinates and poses, are directly incorporated into the policy~\cite{zhang2026qwen,liu2025citywalker,hirose2025omnivla,xue2026omninav}, or 3D structure is reconstructed from RGB images~\cite{zeng2026janusvln,liu2026span}. In some cases, additional geometric sensors fall outside the VLA policy. For example, NaVILA~\cite{cheng2025navila} leverages LiDAR readings for its locomotion controller, while the VLA policy itself relies only on visual input. However, to the best of our knowledge, no prior research has used non-geometric physical quantities as input to VLA navigation.

\subsection{Visual Prompting}
Overlaying marks on input images has proven effective for guiding a model's interpretation. BYOVLA inpaints a task-irrelevant region to recover the nominal behavior under visual distraction~\cite{hancock2025run}, and Visual Attentive Prompting (VAP) applies a tint to the personalized target object~\cite{lee2026bring}. For navigation, waypoint prediction can be enhanced by placing markers on a navigable region~\cite{chen2025affordances}, grounding goal pixels~\cite{bao2026goal2pixel}, or, in concurrent work, segmenting traversable (green) and non-traversable (red) regions~\cite{szvoren2026green}. These visual prompts emphasize, remove, or label information already present in input images. In contrast, we inject visual context entirely absent from the scene: a moving virtual obstacle.

\section{Methodology}

\begin{figure}[t]
\centering
\includegraphics[width=\columnwidth]{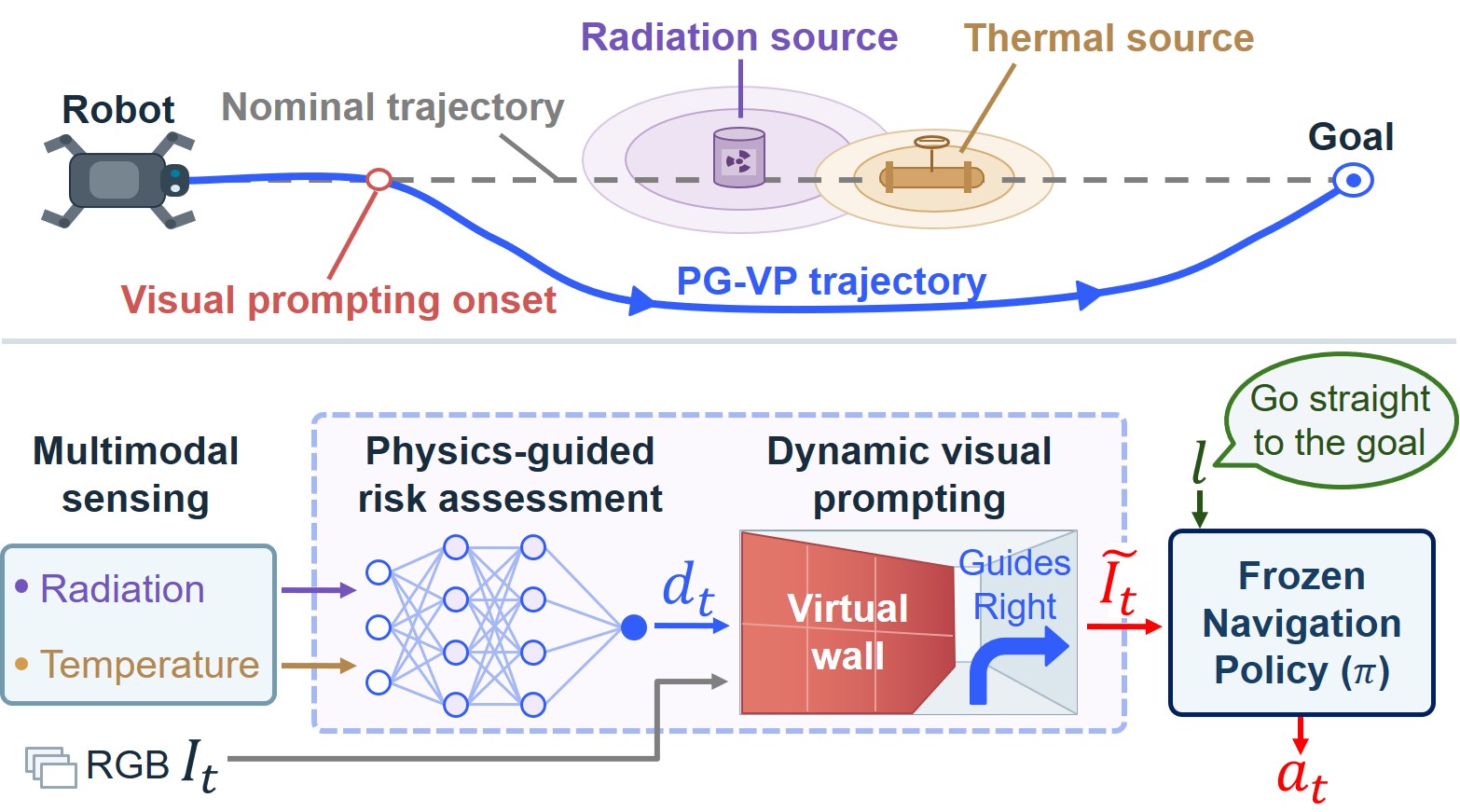}
\caption{Overview of PG-VP framework.}
\label{fig:overall}
\end{figure}

PG-VP detects non-geometric physical quantities using an integrated multimodal perception system (Section~\ref{subsec:integrated_robotic_multimodal_system}), assesses their risk (Section~\ref{subsec:pg-ra}), and guides a navigation policy to naturally avoid proximal invisible hazards (Section~\ref{subsec:visual_prompting}). The policy $\pi$ maps an RGB observation $\mathbf{I}_t$ and a language instruction $\ell$ to an action $a_t \sim \pi(\cdot \mid \mathbf{I}_t, \ell)$. PG-VP overlays a virtual obstacle in the direction $d_t$, converting $\mathbf{I}_t$ into $\tilde{\mathbf{I}}_t$. The action is then inferred as $a_t \sim \pi(\cdot \mid \tilde{\mathbf{I}}_t, \ell)$, while the policy weights and language instruction remain unchanged. When no hazard is detected, $\tilde{\mathbf{I}}_t=\mathbf{I}_t$. Figure~\ref{fig:overall} illustrates the overall flow.

\subsection{Integrated Robotic Multimodal Perception System}
\label{subsec:integrated_robotic_multimodal_system}

Identifying proximal invisible hazards requires auxiliary sensing beyond RGB vision. Our multimodal perception system combines RGB, radiation, and thermal sensing modalities, with radiation and temperature serving as the target non-geometric physical quantities. The vision system consists of a front-facing wide-angle ELP global-shutter camera with a $120^\circ$ horizontal field of view (HFOV) and two webcams with a $69^\circ$ HFOV. The two webcams are yawed by $\pm90^\circ$ relative to the robot's forward direction to provide left and right views. For radiation detection, we use a RadEye SPRD-ER, which provides point-wise gamma-ray measurements at the detector's location. For temperature detection, we use two FLIR Lepton 3.5 radiometric thermal camera modules, each mounted on a PureThermal Mini Pro JST-SR carrier board. To extend horizontal thermal coverage, we mount the two sensors side by side and yaw them $15^\circ$ outward from the robot's forward direction. With $57^\circ$ HFOV for each sensor, this configuration provides a nominal combined HFOV of $87^\circ$, with a $27^\circ$ overlap in the central region.

\begin{figure}[t]
\centering
\includegraphics[width=\columnwidth]{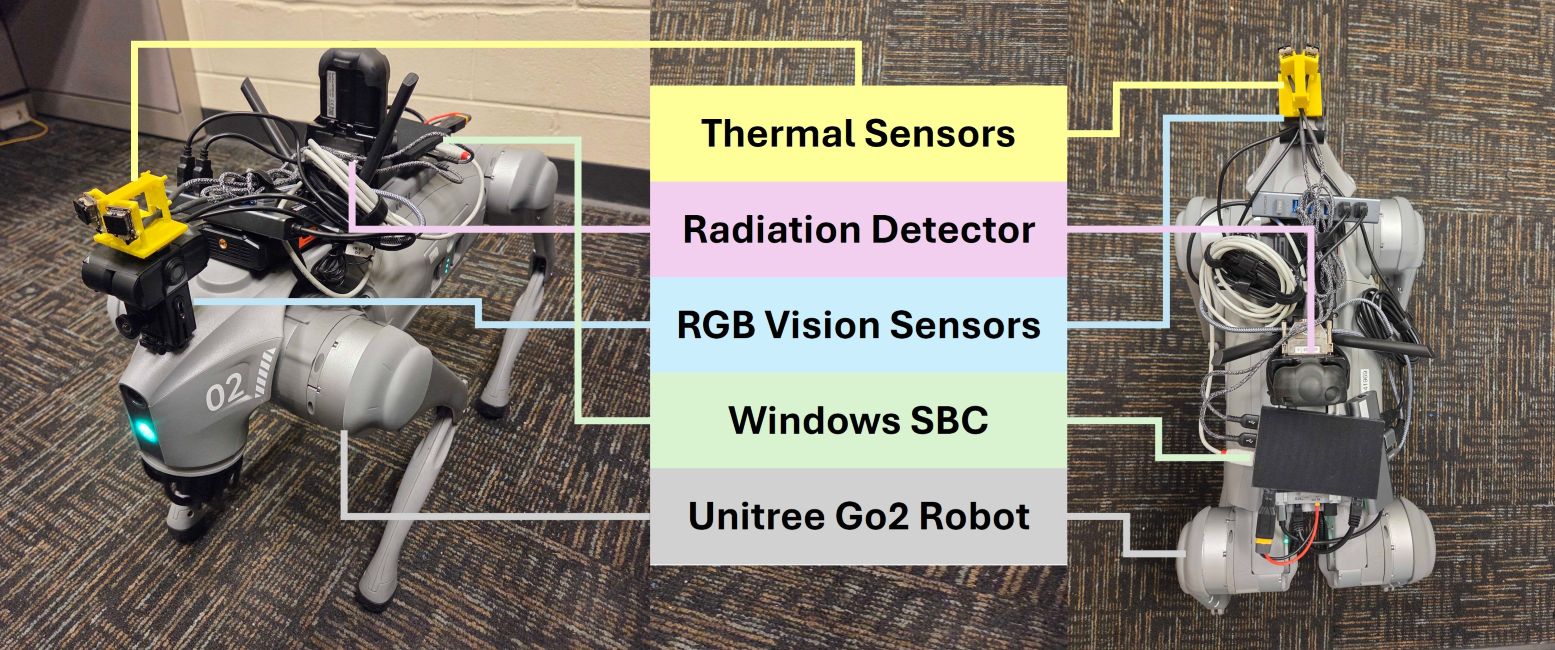}
\caption{Hardware configuration of the robotic multimodal perception system.}
\label{fig:hardware}
\end{figure}

All sensors are mounted on a Unitree Go2 EDU quadruped robot. The RGB cameras are connected directly to the robot's onboard computer, a Jetson Orin Nano. Because of hardware compatibility constraints, the radiation and thermal sensors connect to an external Windows-based single-board computer (SBC). The latest radiation and thermal measurements are transmitted to a Redis server running on the robot's onboard computer, making them available in real time. This configuration is inspired by previous studies of Son~\textit{et al.}~\cite{son2026physics,son2025robot}. Figure~\ref{fig:hardware} shows the sensor placement and data flow of the proposed robotic multimodal perception system.

\subsection{Physics-Guided Risk Assessment}
\label{subsec:pg-ra}

We develop separate risk assessment (RA) models for radiation and temperature. Each RA model operates independently of the navigation policy and is designed around the sensor characteristics and configuration. The models do not explicitly quantify risk; instead, they identify nearby high-risk situations where the robot should take a detour. Each RA model outputs \texttt{left}, \texttt{right}, or \texttt{normal}, depending on the estimated direction of the proximal hazard. The RA models run asynchronously on the robot's onboard computer, and their latest outputs are written to Redis.

\subsubsection{Temperature}
The temperature RA model relies on the hardware configuration, particularly the mounting geometry of the two thermal cameras. As described in Section~\ref{subsec:integrated_robotic_multimodal_system}, the two cameras are mounted side by side and yawed outward by $15^\circ$ from the robot's forward direction. Based on this configuration, the outer one-third of each thermal image is assigned to a side region, while the remaining two-thirds is assigned to a front region. This partition produces four regions: left-side ($\Omega_{\mathrm{LS}}$), left-front ($\Omega_{\mathrm{LF}}$), right-front ($\Omega_{\mathrm{RF}}$), and right-side ($\Omega_{\mathrm{RS}}$).

For each measurement step $t$, the maximum temperature within each region is computed as $T_t^{k}=\max_{\mathbf{p}\in\Omega_k}\mathcal{T}t(\mathbf{p})$, where $k\in\{\mathrm{LS},\mathrm{LF},\mathrm{RF},\mathrm{RS}\}$ and $\mathcal{T}t(\mathbf{p})$ denotes the temperature measured at pixel $\mathbf{p}$. A thermal hazard is classified as being on the left when $T_t^{\mathrm{LS}}
\geq
\max\left(
T_t^{\mathrm{LF}},
T_t^{\mathrm{RF}},
T_t^{\mathrm{RS}}
\right)+\Delta T_{\text{th}}$
and on the right when $T_t^{\mathrm{RS}}
\geq
\max\left(
T_t^{\mathrm{LS}},
T_t^{\mathrm{LF}},
T_t^{\mathrm{RF}}
\right)+\Delta T_{\text{th}}$
, where $\Delta T_{\text{th}}$ is the temperature-difference threshold. A directional output (\texttt{left} or \texttt{right}) is confirmed only if the same classification is maintained for 0.5 seconds over consecutive steps. Otherwise, the temperature RA model outputs \texttt{normal}.

\subsubsection{Radiation}
Unlike the temperature RA model, the radiation RA model relies primarily on algorithmic processing and adopts a physics-informed machine learning (PIML) approach. We developed a two-stage radiation RA model that first assesses radiation risk (risk model) and then estimates the left/right avoidance direction (direction model) from sequences of robot positions, $\mathbf{p}_t=(x_t, y_t)$, and gamma-ray counts per second (CPS). Progressive densification interpolates the available samples to match the required input size when samples are insufficient in the initial input interval. The risk model uses 16 samples: 16 relative log-CPS values, 16 displacement magnitudes, and one flux-heading slope, yielding a 33-dimensional input. When the estimated risk exceeds $0.3$, the direction model is activated with 30 samples, consisting of 30 relative log-CPS values, 60 planar displacement components $(\Delta x, \Delta y)$, 60 heading-change components $(\sin{\Delta \theta}, \cos{\Delta \theta})$, and one flux-heading slope, yielding a 151-dimensional input. The heading is approximated from consecutive robot positions, and the flux-heading slope is obtained by regressing relative log-CPS changes against the corresponding heading changes.

The training data is generated from a $10\mathrm{m}\times10\mathrm{m}$ gamma radiation field computed with OpenMC across 25 arbitrary obstacle environments, using a $100 \times 100$ regular mesh with $0.1\mathrm{m}$ resolution. We retrieve gamma flux sequences from randomly generated robot trajectories in the radiation field, reflecting obstacle-induced attenuation and scattering. In addition, to mirror the sensor condition and counting noise, we randomly configure the flux-to-CPS scaling factor $\kappa\sim\operatorname{LogUniform}(10^2, 10^{10})$, background count rate $b\sim\operatorname{Uniform}(5, 50)$ CPS, and integration time $\tau\sim\operatorname{Uniform}(0.5, 3.0)$. The gamma count, $N_t$, can be modeled following Poisson noise $N_t\sim\operatorname{Poisson}[(\kappa\cdot q(\mathbf{p}_t)+b)\tau]$ where $q(\mathbf{p}_t)$ denotes gamma flux at the point $\mathbf{p}_t$. Then, the modeled input CPS, $c_t$, can be calculated as $c_t = N_t/\tau$. The number of training sample windows for the risk model is 1,051,106, and for the direction model is 911,106. 

Both the risk model and the direction model use multi-layer perceptrons (MLPs). The risk model uses layers of size $(33, 128, 128, 64, 1)$ and uses a class-weighted binary cross-entropy loss. It outputs a float between 0 and 1. The direction model has shared trunk layers with sizes $(151, 128, 128, 64)$, followed by direction head layers with sizes $(64, 32, 3)$ and physics auxiliary head with sizes $(64, 32, 4)$. The direction head outputs the estimated direction probabilities $\hat{\mathbf{y}}_{dir}$ as a 3-dimensional vector, and the physics auxiliary head outputs a vector including the estimated source offset $(\hat{\Delta x_s}, \hat{\Delta y_s})$, flux-to-CPS scaling factor $\hat{\kappa}$, and background count rate $\hat{b}$ (total 4-dimensional output). The direction model leverages a soft-voting ensemble that averages the softmax probabilities of 12 models trained with distinct random seeds. The choice of 12 is inspired by the number of parallel inference models used in the previous PIML-based radiation perception study~\cite{son2026physics}. The loss function, $\mathcal{L}_{dir}$, combines two terms: (i) class-weighted cross-entropy loss, $\mathcal{L}_{ce}$, that directly measures loss between the estimated direction $\hat{\mathbf{y}}_{dir}$ and ground truth direction $\mathbf{y}_{dir}$; (ii) physics-aware auxiliary loss, $\mathcal{L}_{physics}$, that measures the error between the reconstructed CPS, $\hat{c}_t$, and ground truth $c_t$. Equation~\eqref{eq:dir_loss} represents the combined direction model loss.
\begin{equation}
    \mathcal{L}_{dir} = \mathcal{L}_{ce}(\hat{\mathbf{y}}_{dir}, \mathbf{y}_{dir}) + 0.1\mathcal{L}_{physics}(\hat{c}_t, c_t)
    \label{eq:dir_loss}
\end{equation}
The physics-aware auxiliary loss is defined as equation~\eqref{eq:phy_loss}.
\begin{equation}
    \mathcal{L}_{physics} = \frac{1}{30}\sum_{t=1}^{30} |\log(\hat{c}_t+1) - \log(c_t+1)|
    \label{eq:phy_loss}
\end{equation}
Equation~\eqref{eq:c_hat} reconstructs the CPS using the inverse-square law, where $\hat{r}_t(\hat{\Delta x_s}, \hat{\Delta y_s})$ denotes the estimated source-to-detector distance and $r_{\text{det}}$ is the detector radius, $2.82 \mathrm{cm}$.
\begin{equation}
    \hat{c}_t = \frac{\hat{\kappa}}{4 \pi (100\cdot\hat{r}_t(\hat{\Delta x_s}, \hat{\Delta y_s}) + r_{\text{det}})^2} + \hat{b}
    \label{eq:c_hat}
\end{equation}

The physics-guided auxiliary loss encourages the CPS values reconstructed from the estimated source location and intensity to match the observed measurements. This formulation is inspired by Son~\textit{et al.}~\cite{son2026physics}. However, unlike Son's method, the proposed model determines the avoidance direction through a single forward pass of a pre-trained ensemble model. The two-stage radiation RA model achieves 85.7\% accuracy across 14 real-robot short-trajectory scenarios we collected, and 77.4\% accuracy on three real-robot radiation source localization (RSL) scenarios from Son's research~\cite{son2026physics}, demonstrating its acceptable applicability.

\subsection{Dynamic Visual Prompting}
\label{subsec:visual_prompting}

Based on the proximal risk direction estimated by the RA models, we render a virtual obstacle as a visual prompt, maintaining a consistent visual representation across sensing modalities. This representation has two benefits. First, it maps heterogeneous physical risks to a single visual pattern, allowing the same guidance across distinct modalities. If obstacle-like visual patterns fall within the VLA model’s training data distribution, it can interpret the visual prompt without additional modality-specific training. Second, directly overlaying radiation or temperature fields may be ambiguous and require additional semantic interpretation. In contrast, a virtual obstacle is more straightforward because it directly indicates the region that the robot should avoid.

The virtual obstacle is rendered as a dynamic curved wall. The wall gradually grows and shrinks across successive frames. Let $k$ be the relative frame index from the start, and $N_g$ and $N_v$ be the numbers of growth and vanishing frames, respectively. The horizontal wall-length ratio $r_w(k)$ is defined as a nonlinear function of $k$, as shown in equation \eqref{eq:guidance_strength}. Here, $S_\text{guide}$ denotes the guidance strength, defined as the maximum horizontal wall-length ratio.
\begin{equation} 
    r_w(k) = 
    \begin{cases} 
    S_\text{guide} \sqrt{\dfrac{k}{N_g}}, & \text{if } 0 < k \leq N_g \\ 
    S_\text{guide} \sqrt{\dfrac{N_v-(k-N_g)}{N_v}}, & \text{if } N_g < k \leq N_g+N_v
    \\ 0, & \text{otherwise}
    \end{cases} 
    \label{eq:guidance_strength} 
\end{equation}

Let $W$ and $H$ denote the width and height of the front-view image, respectively. The dynamic virtual wall extends inward from either lateral edge of the image. Its horizontal length $l_w(k)$ is calculated as $l_w(k)=r_w(k) W$. Since $S_\text{guide}$ is the maximum $r_w(k)$, the maximum $l_w(k)$ is $S_\text{guide} W$.

Let the origin of the image coordinate system be located at the upper-left corner. For a virtual wall generated from the left edge, the wall region is bounded by the left image edge, an inner vertical boundary at $x=l_w(k)$, and two quadratic B\'ezier curves, $\mathbf{B}(u) = (1-u)^2\mathbf{P}_0 + 2(1-u)u\mathbf{P}_1 + u^2\mathbf{P}_2$, forming the upper and lower boundaries, where $\mathbf{P}_0$ and $\mathbf{P}_2$ are the endpoints, $\mathbf{P}_1$ is the control point, and $u\in[0,1]$.

For the upper curve, the endpoints are $\mathbf{P}_0^u=(0,0.1H)$ and $\mathbf{P}_2^u=(l_w(k),0.3H)$, and the control point is $\mathbf{P}_1^u=(0.06l_w(k),0.2H)$. For the lower curve, the endpoints are $\mathbf{P}_0^b=(d_b(k),H)$ and $\mathbf{P}_2^b=(l_w(k),0.8H)$, and the control point is $\mathbf{P}_1^b=(d_b(k)+0.06l_w(k),0.9H)$. The bottom offset $d_b(k)$ increases linearly from $0$ to $0.2W$ during the growth frames and decreases to $0$ during the vanishing frames. While $\mathbf{P}_0^u$ remains fixed, $\mathbf{P}_0^b$ moves horizontally. This makes the wall appear to rotate from the side into the front view and then move back, rather than simply expanding horizontally. The right virtual wall is generated by mirroring the left virtual wall horizontally.

The wall region is filled with a concrete-like texture generated from Gaussian-noised pixels with a mean of 80 and a standard deviation of 15. Directional shading creates a natural sense of depth, with texture brightness increasing from $50\%$ at the inner boundary to $100\%$ at the image edge. The virtual wall is rendered independently for each front-view frame, and the side-view image corresponding to the wall direction is also completely covered with the same texture.

\begin{figure}[t]
    \centering
    \includegraphics[width=\columnwidth]{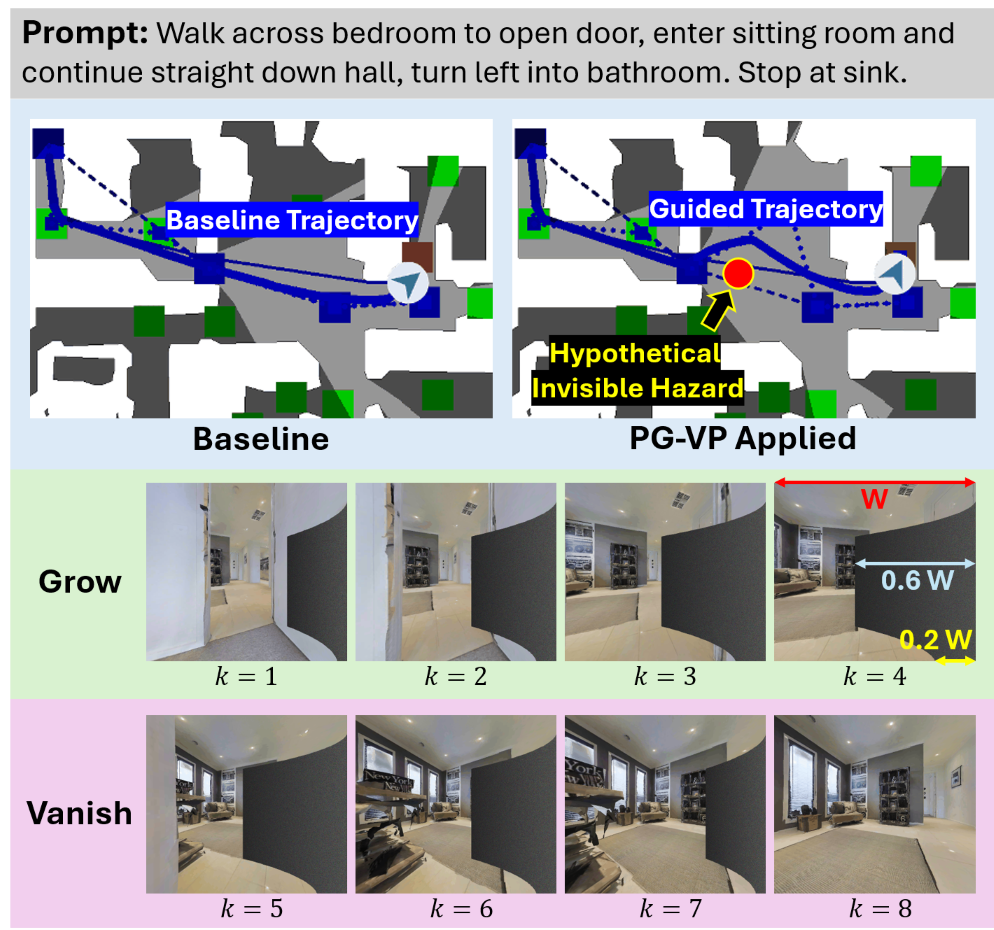}
    \caption{Dynamic visual prompting process in R2R-CE val-unseen sample. (Right wall, $S_{\text{guide}}=0.6$, $N_g = N_v = 4$)}
    \label{fig:pgvp_visualization}
\end{figure}

Figure~\ref{fig:pgvp_visualization} shows the dynamic visual prompting process during the growth and vanishing phases and its effect on the navigation trajectory. It presents an R2R-CE val-unseen sample with a virtual wall generated from the right side of the image, where $S_\text{guide}=0.6$ and $N_g=N_v=4$. To evaluate PG-VP's direct effect, we introduce a hypothetical invisible hazard that bypasses the RA stage and applies only dynamic visual prompting. Compared with the baseline trajectory, the PG-VP-guided trajectory naturally detours around the hypothetical invisible hazard. At $k=4$, $r_w(k)$ reaches its maximum value of $S_\text{guide}$, resulting in a wall length of $l_w(4)=S_\text{guide} W=0.6W$ and a bottom offset of $d_b(4)=0.2W$.

\section{Results}
\subsection{Trajectory Steering on VLN Benchmarks}
We evaluated the trajectory-guidance effectiveness of PG-VP on the val-unseen splits of R2R-CE and RxR-CE~\cite{Matterport3D,anderson2018vision,ku2020room,krantz_vlnce_2020}. The underlying assumption is that an invisible hazard exists in the navigation scenario, as illustrated in Fig.~\ref{fig:pgvp_visualization}, with the virtual wall overlaid at the appropriate location and time step. We first measure the total number of frames in a scenario without PG-VP, then apply visual prompting to a randomly chosen start index between 25\% and 75\% of the total. The number of PG-VP frames is set to $N_g = N_v = 4$, and the guidance strength $S_\text{guide}$ is varied across 0.55, 0.60, and 0.65. We use the OmniNav model as the navigation policy~\cite{xue2026omninav}. 

\begin{figure}[t]
\centering
  \subfloat[CGR ($\%$)]{
  \includegraphics[width=.305\columnwidth]{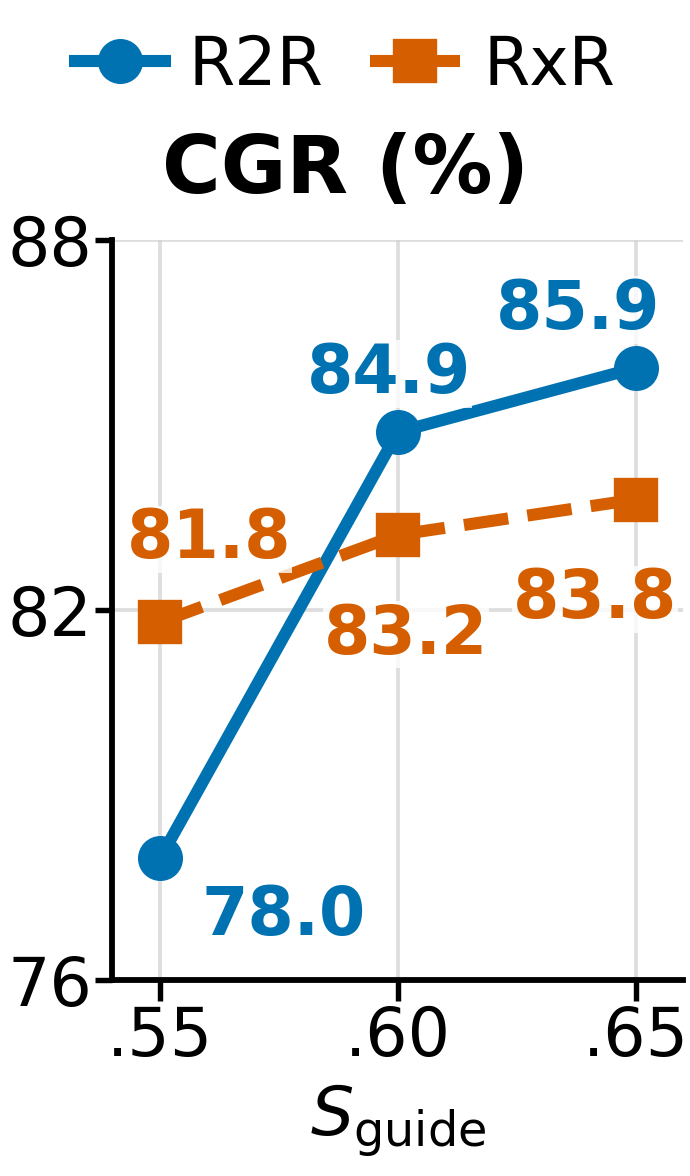}
  \label{subfig:pgvp_correct_guidance}
  }
  \subfloat[Max. $|x_d|$ ($\mathrm{m}$) ]{
  \includegraphics[width=.305\columnwidth]{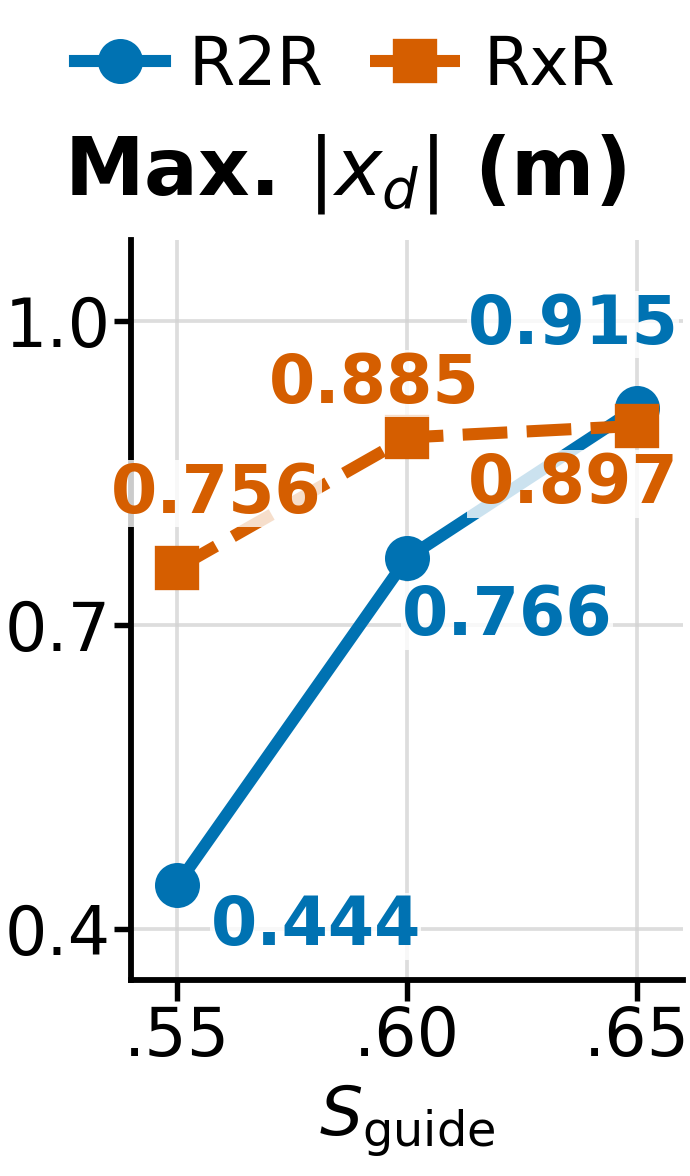}
  \label{subfig:pgvp_post_prompt_max_x}
  }
  \subfloat[SR Drop (pp)]{
  \includegraphics[width=.305\columnwidth]{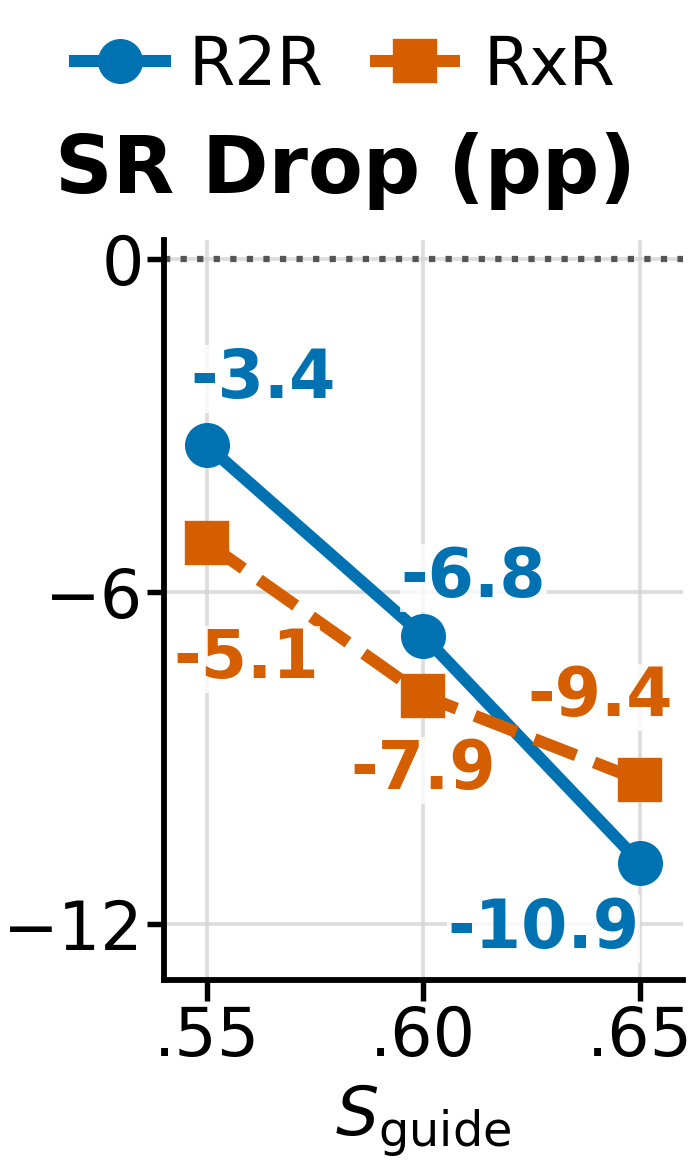}
  \label{subfig:pgvp_sr_change}
  }
  \caption{Trade-off between PG-VP trajectory steering effectiveness and VLN success rate. The baseline model is OmniNav~\cite{xue2026omninav}. (a) Correct-guidance ratio from PG-VP, (b) Maximum displacement in the x-direction from PG-VP, (c) Success rate drop of VLN}
  \label{fig:trade_off}
\end{figure}

The focus here is not on the PG-VP-applied VLA model achieving the best navigation performance, but on how it effectively guides a trajectory to detour around a hypothetical invisible hazard. While PG-VP steers the navigation trajectory appropriately, it reduces the Success Rate (SR) in VLN by introducing new visual context during closed-loop navigation. Figure~\ref{fig:trade_off} shows this performance trade-off in R2R-CE and RxR-CE val-unseen splits. The Correct-Guidance Ratio (CGR) is the percentage of scenarios in which the visual prompt guides the trajectory as intended, and the x-direction displacement, $|x_d|$, measures the magnitude of the robot's lateral displacement induced by the visual prompt. Note that the maximum $|x_d|$ does not occur within the visual-prompt frames; it occurs in the post-prompt phase, 20 extra frames after the visual-prompt phase, from $k=N_g+N_v+1$ to $k=N_g+N_v+20$. That means PG-VP affects the navigation trajectory even after the visual prompting phase.

In Fig.~\ref{subfig:pgvp_correct_guidance}, CGR and $S_\text{guide}$ are positively correlated, indicating that greater guidance strength increases the likelihood of correctly steering the trajectory. Similarly, in Fig.~\ref{subfig:pgvp_post_prompt_max_x}, maximum $|x_d|$ and $S_\text{guide}$ are positively correlated, indicating that greater guidance strength produces larger lateral displacement. Both figures show that guidance strength controls trajectory steering as intended. However, guidance strength comes with a trade-off: it causes an SR drop in VLN. Figure~\ref{subfig:pgvp_sr_change} illustrates the SR drop trend as $S_\text{guide}$ increases; the greater the $S_\text{guide}$, the larger the SR drop. Notably, the slope of the SR drop in R2R-CE becomes steeper, while the slopes of CGR and $\max |x_d|$ become substantially less steep in both R2R-CE and RxR-CE over the $S_\text{guide}$ interval of 0.60–0.65 compared with 0.55–0.60. In particular, CGR and $\max |x_d|$ increase significantly on the R2R-CE splits as $S_\text{guide}$ increases from 0.55 to 0.60. This supports choosing $S_\text{guide}=0.60$ as a reasonable balance between navigation success and guidance effectiveness.

\begin{table}[t]
\caption{Navigation performance (\%) on R2R-CE and RxR-CE Val-Unseen splits.}
\label{tab:vln_benchmark}
\centering
\small
\renewcommand{\arraystretch}{1.05}
\begin{tabular*}{\columnwidth}{
    @{\extracolsep{\fill}}lccccc@{}
}
\toprule
\multirow{2}{*}{Method}
& \multirow{2}{*}{Obs.}
& \multicolumn{2}{c}{R2R-CE}
& \multicolumn{2}{c}{RxR-CE} \\
\cmidrule(lr){3-4}
\cmidrule(l){5-6}
& & SR$\uparrow$ & SPL$\uparrow$
  & SR$\uparrow$ & SPL$\uparrow$ \\
\midrule
NaVILA (2025)~\cite{cheng2025navila}
& S.RGB & 54.0 & 49.0 & 49.3 & 44.0 \\

StreamVLN (2025)~\cite{wei2025streamvln}
& S.RGB & 56.4 & 50.2 & 54.4 & 45.4 \\

CorrectNav (2026)~\cite{yu2026correctnav}
& S.RGB & 65.1 & 62.3 & 69.3 & 63.3 \\

JanusVLN (2026)~\cite{zeng2026janusvln}
& S.RGB & 60.5 & 56.8 & 56.2 & 47.5 \\

\addlinespace[1pt]
NavFoM (2026)~\cite{zhang2026embodied}
& Pano. & 61.7 & 55.3 & 64.4 & 56.2 \\

OmniNav (2026)~\cite{xue2026omninav}
& Pano. & 69.5 & 66.1 & 73.6 & 62.0 \\

ABot-N0 (2026)~\cite{chu2026abot}
& Pano. & 66.4 & 63.9 & 69.3 & 60.0 \\

SPAN-Nav (2026)~\cite{liu2026span}
& Pano. & 66.3 & 59.3 & 69.7 & 60.1 \\
\midrule
OmniNav (our run)
& Pano. & 68.7 & 65.6 & 68.8 & 57.8 \\

PG-VP ($S_{\text{guide}}{=}0.55$)
& Pano. & 65.4 & 61.7 & 63.7 & 52.2 \\

PG-VP ($S_{\text{guide}}{=}0.60$)
& Pano. & 61.9 & 58.2 & 61.0 & 49.7 \\

PG-VP ($S_{\text{guide}}{=}0.65$)
& Pano. & 57.8 & 54.3 & 59.4 & 48.0 \\
\bottomrule
\end{tabular*}
\end{table}

In addition, the navigation performance of the OmniNav model with PG-VP ($S_\text{guide} = 0.60$) is acceptable compared to other previous methods. Table~\ref{tab:vln_benchmark} presents the SR and the Success weighted by Path Length (SPL) across various recent navigation methods on the benchmarks. The baseline OmniNav performance we obtained by running the official code differs from the results reported in the original paper~\cite{xue2026omninav}. Even with the performance drop, PG-VP with $S_\text{guide}=0.60$ outperforms most single-camera-based methods except CorrectNav. It also maintains performance comparable to NavFoM, a panoramic camera-based method, although it falls behind other panoramic-view methods. Even so, PG-VP with $S_\text{guide}=0.60$ does not substantially degrade navigation performance relative to other recent methods. Reducing $S_\text{guide}$ to 0.55 improves navigation performance, but the insight from Fig.~\ref{fig:trade_off} suggests that the resulting trajectory guidance is too weak. Increasing $S_\text{guide}$ to 0.65, by contrast, slightly improves guidance effectiveness, but navigation performance drops sharply and is no longer competitive with other recent methods.
\begin{figure}[t]
\centering
  \subfloat[R2R-CE]{
  \includegraphics[width=.473\columnwidth]{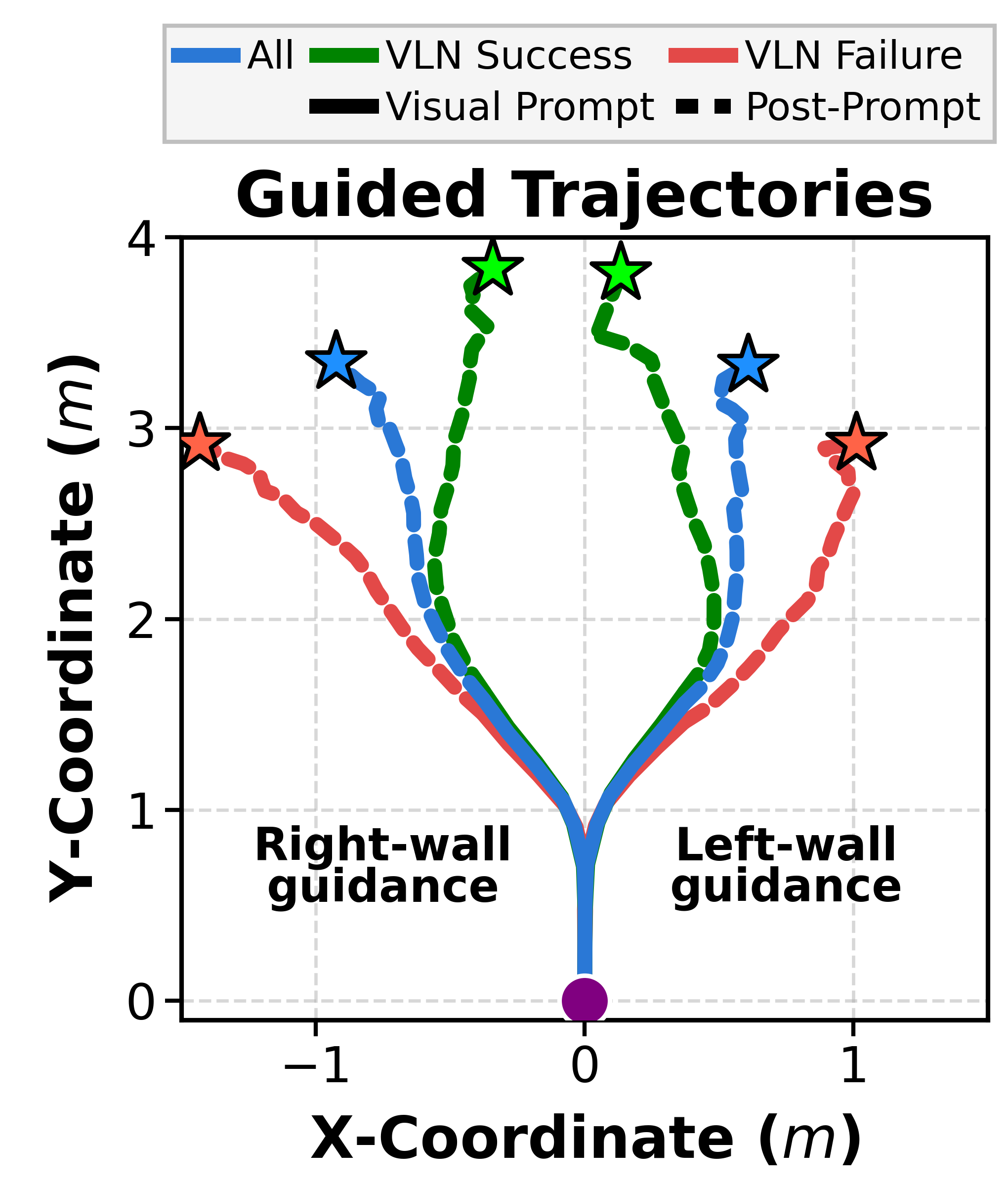}
  \label{subfig:traj_060_r2r}
  }
  \subfloat[RxR-CE]{
  \includegraphics[width=.473\columnwidth]{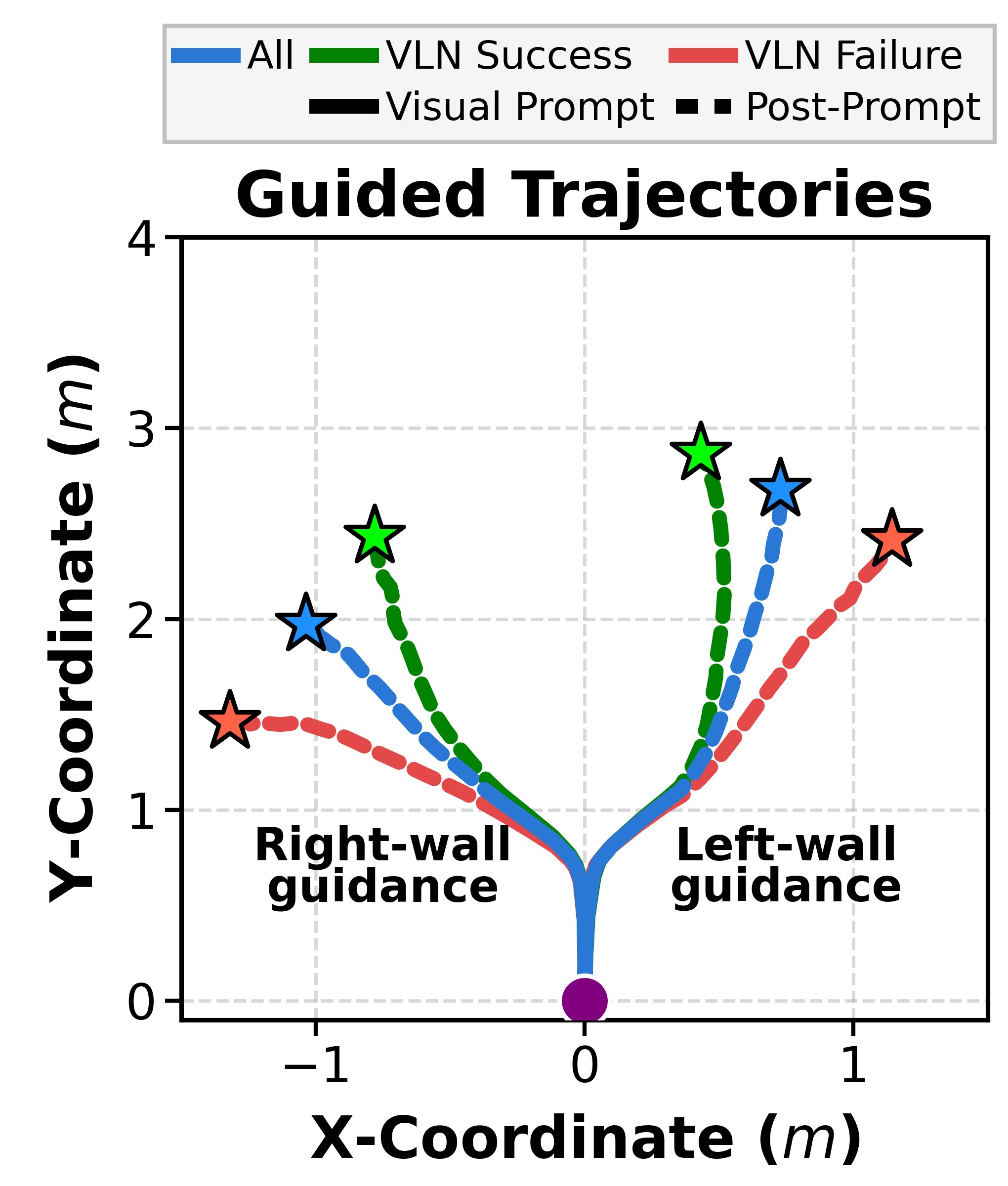}
  \label{subfig:traj_060_rxr}
  }
  \caption{Cartesian average of PG-VP trajectories on the R2R-CE and RxR-CE val-unseen splits. ($S_{\text{guide}}=0.6$, $N_g = N_v = 4$)}
  \label{fig:trajectory}
\end{figure}

Figure~\ref{fig:trajectory} shows the Cartesian average of PG-VP ($S_\text{guide}=0.60$, $N_g=N_v=4$) trajectories on the R2R-CE and RxR-CE val-unseen splits, presenting the qualitative result. The solid lines show the agent's trajectories during the visual prompting phase, when the virtual obstacle is overlaid, while the dashed lines show its trajectories during the post-prompting phase, which covers the following 20 frames. Overall (blue line), dynamic visual prompting guides the trajectories as intended, with trajectories tending to continue in the intended steering direction even after the visual prompting phase ends. Notably, when navigation missions are successfully completed (green lines), trajectories in the post-prompting phase tend to return to the original heading after briefly continuing in the steered direction. However, trajectories keep continuing in the steered direction when navigation failures occur (red line).

\subsection{Experimental Validation}
We experimentally validated PG-VP's low-risk navigation trajectories by fully integrating the system: a physical robot (Section~\ref{subsec:integrated_robotic_multimodal_system}), the physics-guided RA models (Section~\ref{subsec:pg-ra}), and dynamic visual prompting (Section~\ref{subsec:visual_prompting}). At each time step, the VLA model runs on a remote server and communicates with the robot via HTTP, while the RA models and dynamic visual prompting run onboard. Four distinct scenarios are configured: Scenarios A and B test a thermal hazard with $\Delta T_{\text{th}}=30$, and scenarios C and D test a radiation hazard. 

\begin{figure}[t]
    \centering
    \includegraphics[width=\columnwidth]{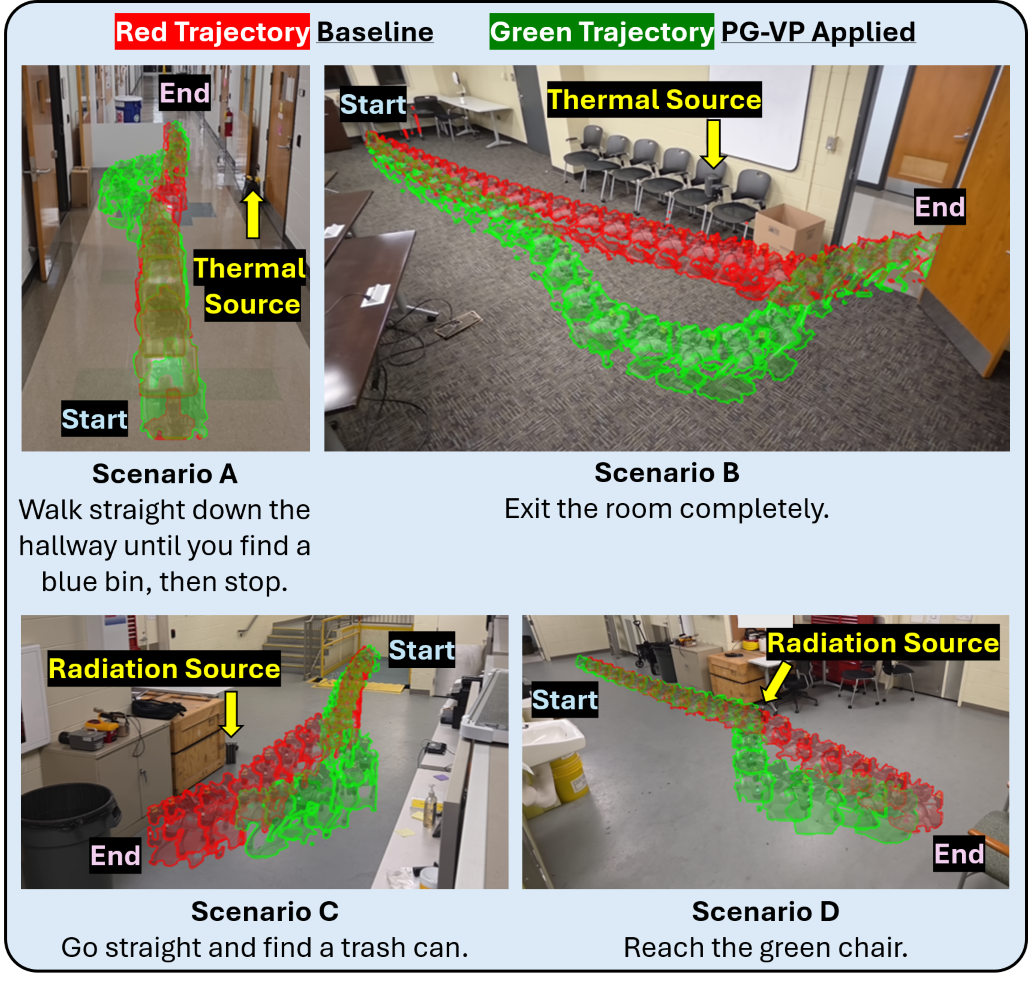}
    \caption{Real robot demonstration ($S_\text{guide}=0.6$, $N_g=N_v=4$, $\Delta T_\text{th}=30$). A: Fine-grained instruction, B: Coarse instruction, C: Exploration task, D: Reaching task.}
    \label{fig:robot_demo}
\end{figure}

Figure~\ref{fig:robot_demo} illustrates navigation language prompts, hazard locations, and robot trajectories with and without PG-VP, when $S_\text{guide}=0.6$ and $N_g=N_v=4$. The language prompt is a fine-grained instruction (Scenario A), a coarse instruction (Scenario B), an exploration task (Scenario C), or a reaching task (Scenario D). Safety concerns restrict the use of highly dangerous hazards: Scenarios A and B use a heat gun and a room heater as thermal hazards, respectively, and Scenarios C and D use nine-bundled uranium rods as a radiation hazard. For all scenarios, the navigation policy successfully achieves the goal, and the PG-VP-applied trajectory (green) effectively detours around a thermal or radiation hazard compared to the baseline trajectory (red). Note that if the visual appearance matches these scenarios but no actual thermal or radiation hazards exist, the PG-VP trajectory will follow the baseline; for example, turning off the heat gun or room heater, or using non-radioactive iron rods.

We define hazard-specific safety metrics based on each sensor's measurement characteristics. Thermal cameras measure the temperature of object surfaces rather than the ambient temperature at the robot’s location. We therefore use heat-source proximity (distance to the heat source) as a thermal safety metric. In contrast, the radiation detector measures gamma-ray counts at the robot’s location. We thus use the net count rate (background-subtracted gamma-ray CPS) to quantify the radiation levels encountered along the trajectory. The background radiation is 31 CPS in scenarios C and D.

\begin{table}[t]
\centering
\caption{Safety metrics comparison between the baseline and PG-VP.}
\label{tab:physical_risk_comparison}
\small
\setlength{\tabcolsep}{2.5pt}
\renewcommand{\arraystretch}{1.05}
\begin{tabular}{@{}ccccccc@{}}
\toprule
\multirow{2}{*}{Method}
& \multicolumn{3}{c}{Heat-source proximity (m) $\uparrow$}
& \multicolumn{3}{c}{Net count rate (CPS) $\downarrow$} \\
\cmidrule(lr){2-4}\cmidrule(l){5-7}
& Scn. & Min. & L-10$\%$ Avg.
& Scn. & Peak & H-10$\%$ Avg. \\
\midrule
Baseline & \multirow{2}{*}{A} & 1.06 & 1.09
         & \multirow{2}{*}{C} & 93.00 & 92.00 \\
PG-VP    &                    & \textbf{1.53} & \textbf{1.55}
         &                    & \textbf{46.00} & \textbf{44.28} \\
\midrule
Baseline & \multirow{2}{*}{B} & 0.79 & 0.85
         & \multirow{2}{*}{D} & 149.00 & 147.81 \\
PG-VP    &                    & \textbf{1.54} & \textbf{1.57}
         &                    & \textbf{137.00} & \textbf{128.14} \\
\bottomrule
\end{tabular}
\end{table}

\begin{figure}[t]
\centering
  \subfloat[Scenario A]{
  \includegraphics[width=.473\columnwidth]{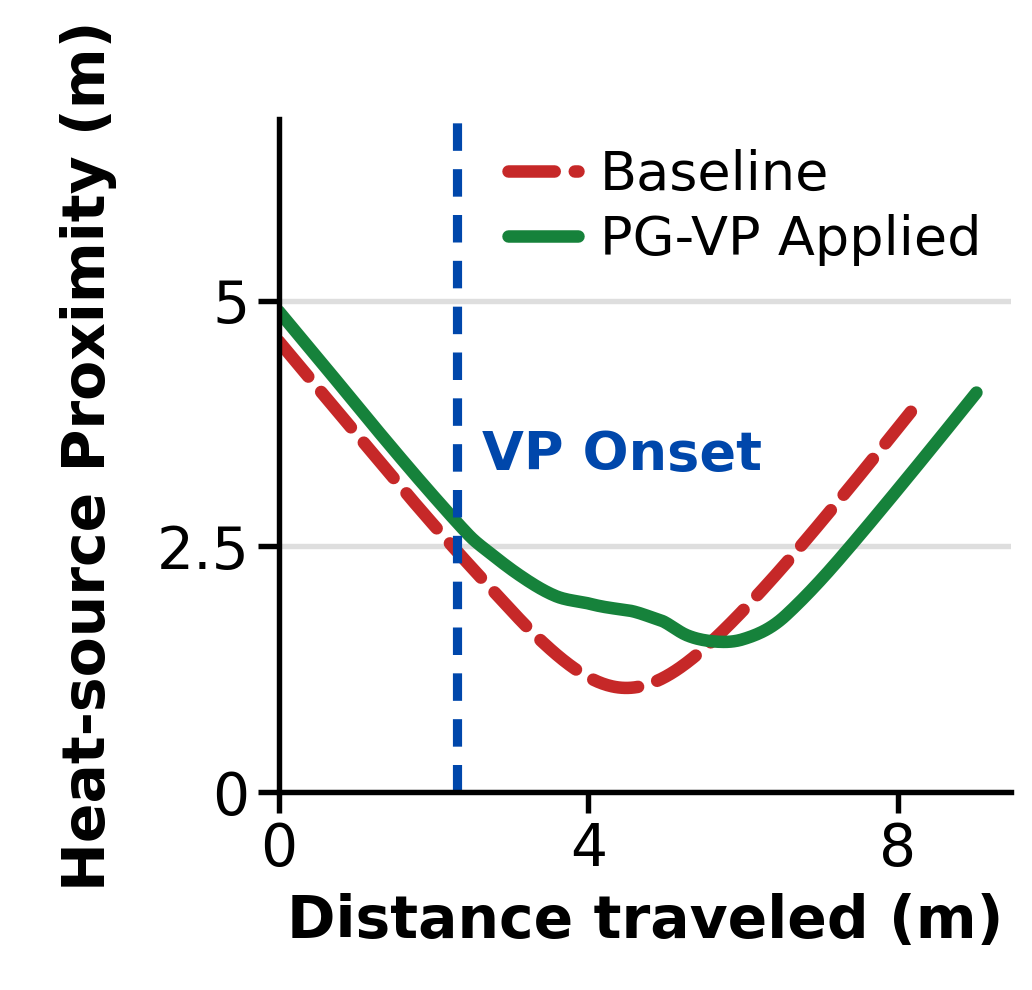}
  \label{subfig:scn_a}
  }
  \subfloat[Scenario B]{
  \includegraphics[width=.473\columnwidth]{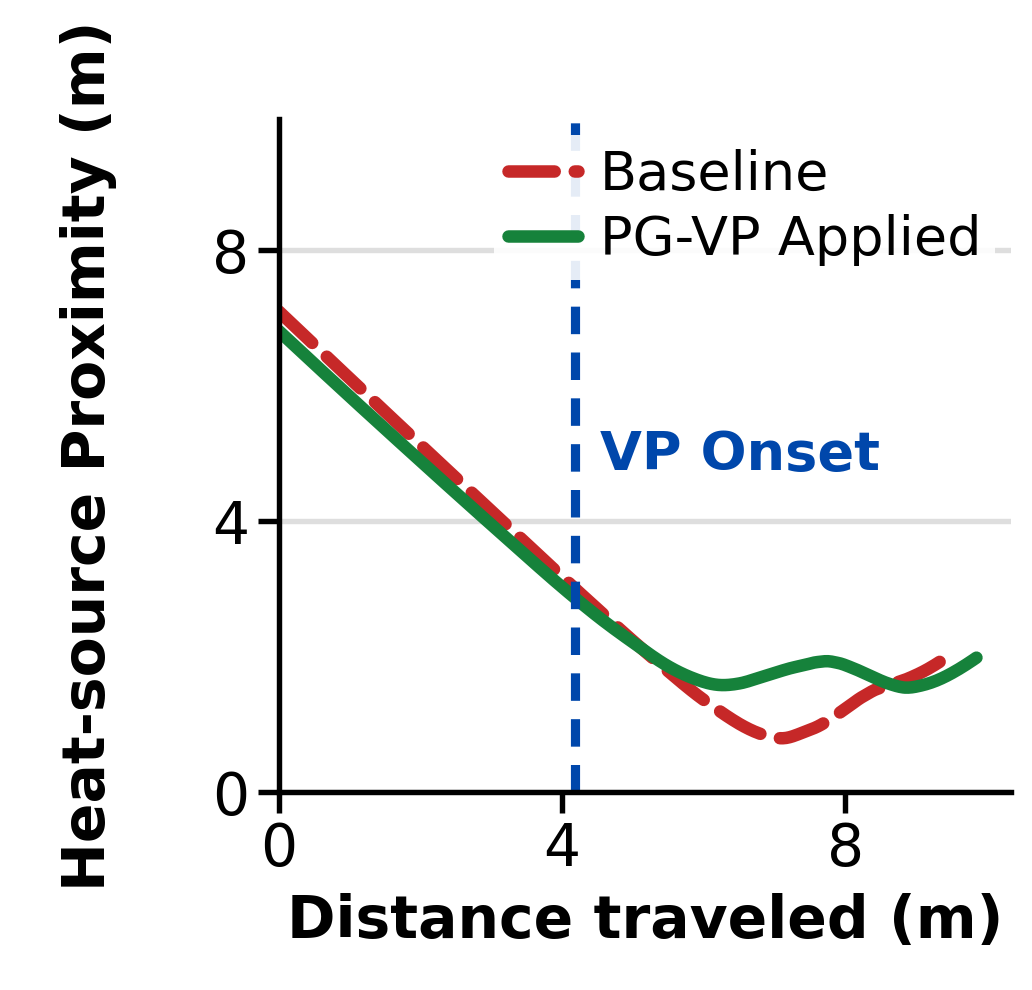}
  \label{subfig:scn_b}
  }
  \vfill
  \subfloat[Scenario C]{
  \includegraphics[width=.473\columnwidth]{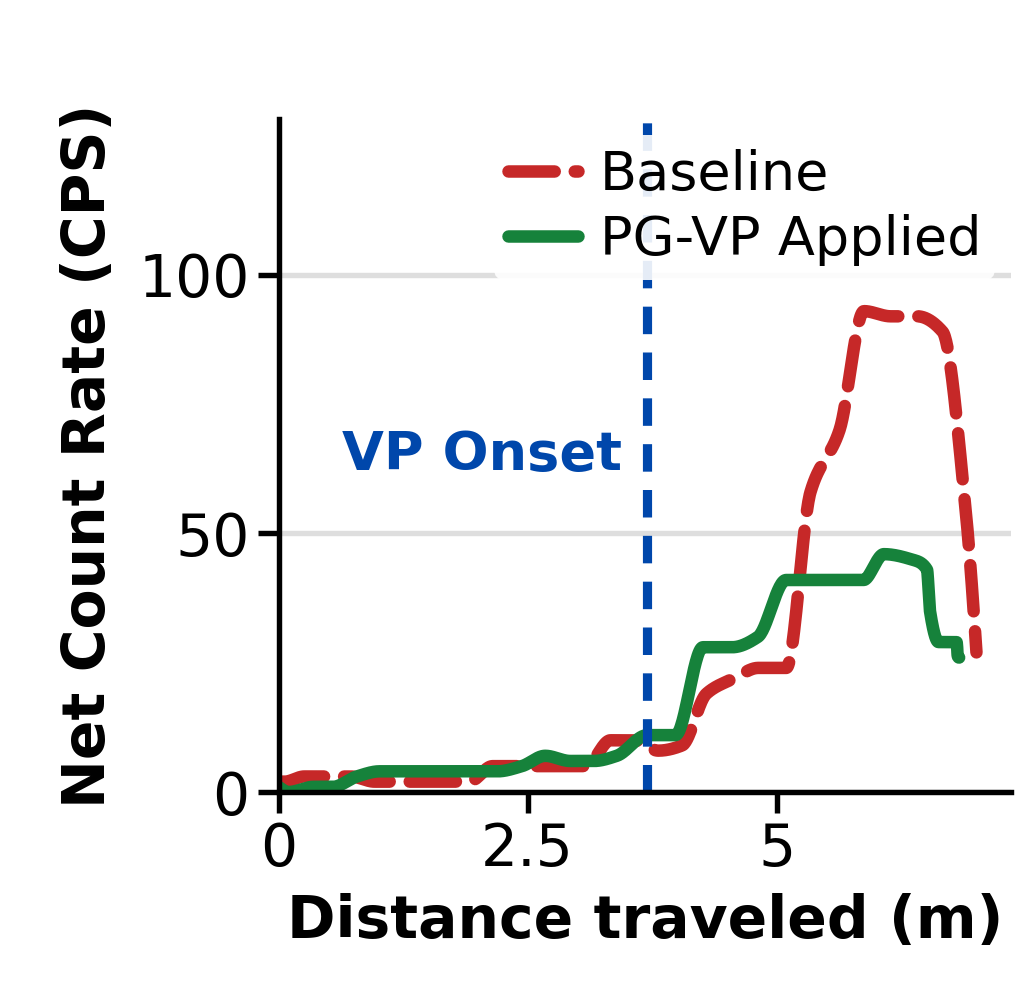}
  \label{subfig:scn_c}
  }
  \subfloat[Scenario D]{
  \includegraphics[width=.473\columnwidth]{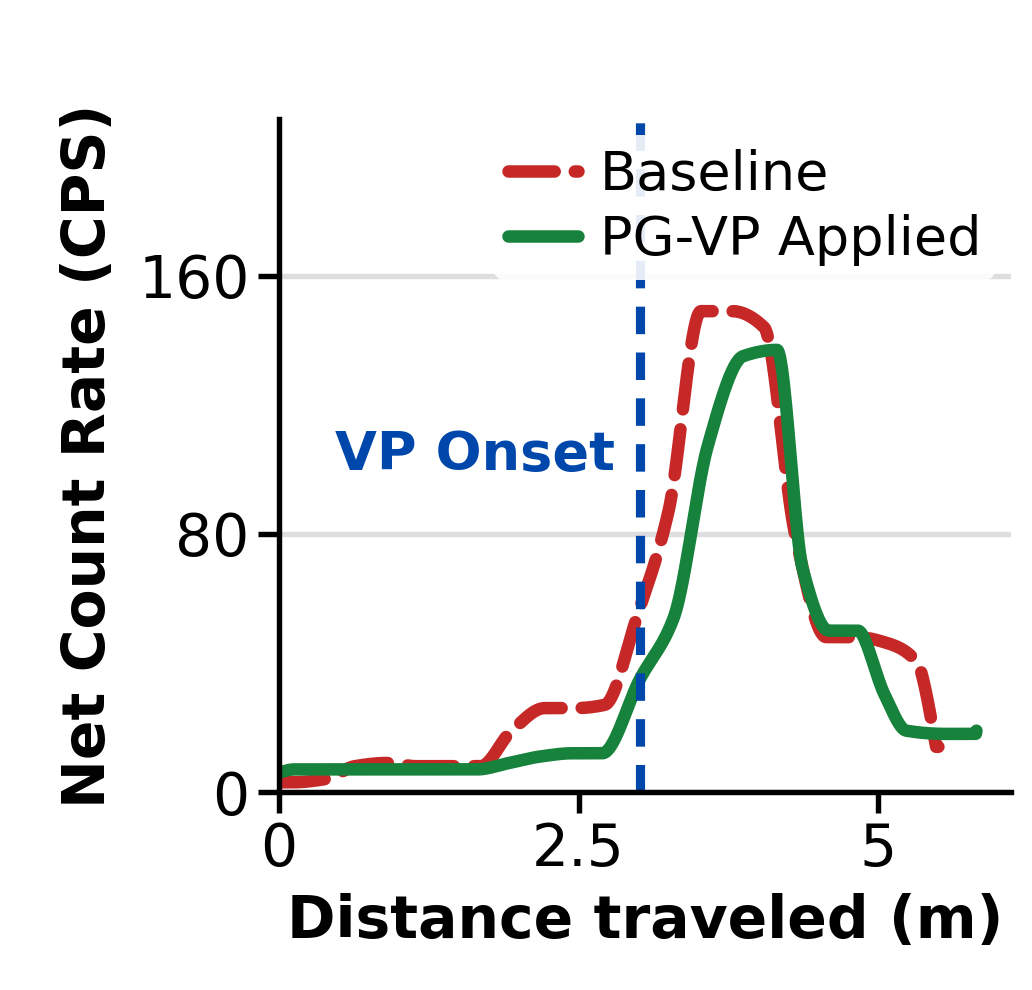}
  \label{subfig:scn_d}
  }
  \caption{Validation of low-risk navigation trajectories from the safety metrics comparison between the baseline and PG-VP-applied model. All are plotted as a function of distance traveled.}
  \label{fig:scn_validation}
\end{figure}

Table~\ref{tab:physical_risk_comparison} compares safety metrics between the baseline and PG-VP-applied model. We compare the extreme value (minimum heat-source proximity or peak net count rate) with the average of the worst 10\% (lowest 10\% or highest 10\%). While the extreme value shows the instantaneous risk, the worst 10\% average quantifies sustained hazard exposure and improves robustness to transient sensor noise. For the extreme value, the PG-VP-applied model outperforms the baseline by 69.64\% for thermal hazard and 29.30\% for radiation hazard. Also, for the average of the worst 10\%, it improves by 63.45\% and 32.59\% for thermal and radiation hazards, respectively.

Figure~\ref{fig:scn_validation} illustrates the safety metrics as a function of the robot's travel distance, along with the onset of the visual prompt. For Scenarios A and B, the baseline heat-source proximity is V-shaped, whereas PG-VP flattens it around the minimum. For Scenarios C and D, the baseline net count rate peaks sharply. While PG-VP nearly eliminates the sharp peak in Scenario C, a peak still remains in Scenario D, but with a lower magnitude and slope. PG-VP onset points can explain why Scenario D fails to eliminate the sharp peak. In Scenarios A, B, and C, RA models trigger the visual prompt well before the risk escalates. In contrast, in Scenario D, the RA model fails to trigger PG-VP early enough because the box occludes the radiation source, as shown in Fig.~\ref{fig:robot_demo}, constraining the net count rate to a constant level. PG-VP is activated immediately after a sharp count-rate surge, delaying hazard detours.

\section{Conclusions}
We proposed a physics-guided visual prompting method that naturally steers the VLA-based navigation trajectory to detour around non-geometric physical hazards in a zero-shot manner. By fusing radiation and thermal data, the risk assessment model determines the avoidance direction. The dynamic visual prompting module then overlays a corresponding virtual wall on the input RGB image to guide the frozen navigation policy's decision. The method is validated in simulation benchmarks and physical experiments. Results show that PG-VP effectively guides navigation trajectories and reduces the invisible risk from actual thermal and radiation sources.

Future work can include a broader range of real-world navigation scenarios, not only in laboratory settings but also in industrial environments, to rigorously validate the integrated PG-VP framework. Another direction is to investigate how PG-VP can achieve modality-specific objectives beyond hazard avoidance, such as localizing radiation sources or finding high-temperature spots. Furthermore, the framework can be extended beyond radiation and thermal sensing to other modalities, such as gas sensing, by designing corresponding risk assessment models while keeping the core navigation policy unchanged.

\bibliographystyle{IEEEtran}
\bibliography{main}

\newpage

\vfill

\end{document}